# Uncertainty and Explainable Analysis of Machine Learning Model for Reconstruction of Sonic Slowness Logs

Hua Wang[1*], Yuqiong Wu[1], Yushun Zhang[1], Fuqiang Lai[2], Zhou Feng[3], Bing Xie[4], Ailin Zhao[4]

1. School of Resources and Environment, University of Electronic Science and Technology of China, Chengdu, Sichuan Province, P.R. China, 611731；

2. School of Petroleum Engineering, Chongqing University of Science and Technology, Chongqing, P.R. China, 401331

3. CNPC Research Institute of Petroleum Exploration and Development

4. Exploration and Development Research Institute, PetroChina Southwest Oil and Gas Field Company, Chengdu, 610041

## Summary

Logs are valuable information for oil and gas fields as they help to determine the lithology of the formations surrounding the borehole and the location and reserves of subsurface oil and gas reservoirs. However, important logs are often missing in horizontal or old wells, which poses a challenge in field applications. To address this issue, conventional methods involve supplementing the missing logs by either combining geological experience and referring data from nearby boreholes or reconstructing them directly using the remaining logs in the same borehole. Nevertheless, there is currently no quantitative evaluation for the quality and rationality of the constructed log. In this paper, we utilize data from the 2020 machine learning competition of the Society of Petrophysicists and Logging Analysts (SPWLA), which aims to predict the missing compressional wave slowness (DTC) and shear wave slowness (DTS) logs using other logs in the same borehole. We employ the natural gradient boosting (NGBoost) algorithm to construct an Ensemble Learning model that can predicate the results as well as their uncertainty. Furthermore, we combine the SHAP (SHapley Additive exPlanations) method to investigate the interpretability of the machine learning model. We compare the performance of the NGBosst model with four other commonly used Ensemble Learning methods, including Random Forest, GBDT, XGBoost, LightGBM. The results show that the NGBoost model performs well in the testing set and can provide a probability distribution for the prediction results. This distribution allows petrophysicists to quantitatively analyze the confidence interval of the constructed log. In addition, the variance of the probability distribution of the predicted log can be used to justify the quality of the constructed log. Using the SHAP explainable machine learning model, we calculate the importance of each input log to the predicted results as well as the coupling relationship among input logs. Our findings reveal that the NGBoost model tends to provide greater slowness prediction results when the neutron porosity (CNC) and gamma ray (GR) are large, which is consistent with the cognition of petrophysical models. Furthermore, the machine learning model can capture the influence of the changing borehole caliper on slowness, where the influence of borehole caliper on slowness is complex and not easy to establish a direct relationship. These findings are in line with the physical principle of borehole acoustics. Finally, by using the explainable machine learning model, we observe that although we did not correct the effect of borehole caliper on the neutron porosity log through preprocessing, the machine learning model assigned a greater importance to the influence of the caliper, achieving the same effect as caliper correction.

[*] Correspoding author, Email: huawang@uestc.edu.cn

Keywords: Missing Log Construction, Ensemble Learning, NGBoost, Quality Control

## 1. Introduction

Ensuring control over a country's oil and gas resources requires reducing exploration cost and increasing production efficiency. One effective means of achieving this is borehole logging, a widely used technique for determining the properties of subsurface reservoirs(Asquith et al. 2004). However, borehole logs may be missed during field application due to several issues such as cost control, tool failure, poor logging conditions, data processing errors, and human factors during data acquisition(Jia et al. 2012). In detail, logs such as natural gamma ray, resistivity, density, and neutron porosity are commonly acquired at relatively low cost. However, more expensive logs such as nuclear magnetic resonance (NMR) and sonic slowness logs are only acquired in a few wells. Tool failure, such as poor contact between the probe and the wellbore or low battery levels, can cause logs to be missed. Poor logging conditions, such as horizontal borehole, can make the measurement difficult, resulting in missing logs. Logging data requires complex processing to obtain the final result(Ellis & Singer, 2007). During processing, errors can occur that result in missing data. These issues can lead to inaccurate interpretation, resulting in suboptimal decisions on hydrocarbon production and management. Therefore, it is essential to address these issues to ensure that borehole logs are obtained accurately and reliably, and that decisions based on these logs are sound.

To address these issues, petrophysicists have several options avaible to them. They can use nearby borehole sections or the other logs from the same section to construct the missing logs(Ghosh 2022). Conventional methods for constructing missing logs involve estimating the missing data points based on statistical techniques such as regression or interpolation(Spacagna et al. 2022). Linear regression is the most commonly used method, which involves finding a linear relationship between the available data points and then using the linear relationship to estimate the missing data points. Another widely used method is interpolation, which involves estimating the missing data points based on the values of nearby data points using mathematical formulas. However, these methods may not be as accurate when the data is complex, or the missing data points are located further away from the available data points. To impove the accuracy, petrophysicsists can also incorporate their local geological knowledge when reconstructing missing logs.

In addition to the conventional methods, advanced machine learning methods are also being employed to reconstruct the missing logs(Choubey & Karmakar 2021; Sircar et al. 2021; Rostamian et al. 2022). Since the 1990s, the logging industry has attempted to achieve automatic interpretation of new logs by using core analysis or expert interpretation results. However, these methods have limited effectiveness due to limitations such as computing capablity, data labels, and intelligent algorithms. With the development of information technology and big data, data-driven methods have been employed in various industries to explore a larger function space within the data, discover potential knowledge from a new perspective, and link data with objectives (Bergen et al. 2019).

In recent years, supervised machine learning algorithms, such as classification and regression problems, have becom increasingly prevalent in logs interpretation(Wang & Zhang 2021). Regression is often used to predict physical parameters from logs. For instrances, Wang et al. (2007) used a BP neural network to predict formation permeability from wireline formation tester data. Li & Misra (2019) used Variational Autoencoders and Long-short Term Memory (LSTM) to construct a model of nuclear magnetic resonance T2 distribution that quantifies the pore size distribution of oil and gas in shale reservoirs. Chopra et al. (2022) predicted porosity from logs in the Volve oilfield in the Norwegian North Sea using a deep neural network. Regression in supervised machine learning method can also be used to reconstruct missing logs. Meshalkin et al. (2020) combined multiple supervised learning algorithms such as AdaBoost, Gradient Boosting, and Extra Trees to construct a rock thermal conductivity prediction model. The model can accurately predict the thermal conductivity curve from the well logs. Chopra et al. (2022) established a deep neural network prediction model for gamma ray in the Volve oil field in the Norwegian North Sea.

However, the machine learning methods have limitations in field applications, with two primary challenges being the uncertainty of the predicted results and the lack of model interpretability.

**(1) Uncertainty analysis:**

There are no "labels" to evaluate the model performance when these machine learnding methods are applied to blind (or new) wells. In addition, there is no way to prove the effectiveness of the reconstructed log except for the expensive coring or re-logging. We need to answer two questions: What is the likelihood that petrophysicists will adopt the model's predicted results? And how can we quantify the uncertainty of the predicted results?

**(2) Model interpretability:**

The current reconstruction of missing logs depends entirely on the petrophysicists' regional geological knowledge. There is no quantitative method to explain why certain logs can be used to reconstruct the missing logs.

They are also common problems in the other applications of machine learning models. To address these limiataions, machine learning methods such as natural gradient boosting tree model (NGBoost) are employed to reconstruct missing logs and provide confidence intervals for predicted results. These methods can help petrophysicists adopt the model's predicted results with more confidence. At the same time, we use the SHAP explainable machine learning model to obtain the importance of each input log to the predicted results and explain the coupling relationship among input logs.

These methods have been applied to ground motion parameters' prediction by Chen & Wang (2022). This paper applies the same idea to the reconstruction of missing well logs. By employing advanced machine learning methods, petrophysicists can improve their ability to accurately and efficiently reconstruct missing well logs, leading to more accurate interpretations and better decisions regarding hydrocarbon production and management.

## 2. Methods

The supervised machine learning method utilized in this study is an ensemble learning method (Dietterich 2002), where multiple individual base learners are combined through a specific strategy to compensate for errors generated by a single learner, resulting in better generalization performance. This approach can be applied to improve machine learning tasks, including classification and regression. The individual learner can be any classification or regression algorithm, such as linear regression models, decision trees, and neural network algorithms.

Decision trees are commonly used in data mining and provide effective learning models for classification tasks that can be represented graphically or by rules. A decision tree consists of a root node, multiple internal nodes, and leaf nodes. Each internal node poses a question-and-answer test for a specific attribute, and each branch below the internal node corresponds to the output of the corresponding question-and-answer test. Each leaf node corresponds to a decision result. Since the size of the tree is independent of the database size, the overall computational complexity of the tree is relatively small. In addition, decision trees are insensitive to the scaling of feature vectors, and the splitting point of the tree is not affected by numerical scaling. As a result, decision trees are suitable for classification and regression of tabular datasets, such as well logs. Different from previous studies on machine learning prediction of well logs (Misra et al. 2019; Gu et al. 2021), this study mainly uses an ensemble learning model with decision trees as individual learners for parameter prediction. We don't need to normalize or scale the well logs.

To improve the generalization performance of ensemble learning, it is important for individual learners to be as independent as possible. There are two categories of ensemble learning based on the construction and generation method of individual learners: parallel and sequential methods.

In parallel ensemble methods, individual base learners are independent of each other, and there is no strong dependency relationship among them. Typically, the entire dataset is divided into multiple subsets to train individual base learners, achieving independence between different learners. Bagging algorithm (Breiman 1996) and Random Forest (RF) (Breiman 2001) are typical representatives of parallel methods. The Bagging algorithm uses bootstrap sampling method with replacement to bring overlapping but different subsets to different base learners. Random Forest is based on the Bagging algorithm with decision trees as base learners and introduces random attribute selection during the training.

In sequential ensemble methods, there is a strong dependency relationship between individual learners, and the output of each base learner will affect the construction of the next base learner. Boosting algorithms are typical representatives of sequential methods, which can boost weak learners into strong learners. It first trains a base weak learner using the initial training set, and then adjusts the distribution of the training labels based on the performance of the base learner to focus on the training labels where the previous base learner made wrong judgments. Then, the next base learner is trained based on the adjusted distribution of labels. Finally, all base learners are combined with different weights.

In this study, we use various sequential ensemble methods to reconstruct the missing log from other logs at the same depth section in the same borehole, including Random Forest (RF) (Breiman 2001), GBDT (Friedman 2001), XGBoost (Chen & Guestrin 2016), LightGBM (Ke et al. 2017), and Natural Gradient Boosting (NGBoost) (Duan et al. 2020) . NGBoost is a new boosting algorithm that incorporates natural gradient in the Riemann domain (Amari 1998) to optimize the learning process along the steepest descent direction rather than the ordinary gradient direction in the Euclidean domain, resulting in more efficient and stable machine learning outcomes. In addition, the NGBoost not only provide the preditiocon but the probability distribution of the predication. Fig. 1 illustrates the relationship between sequential ensemble methods. The base learners used in this study are classification and regression trees (CART).

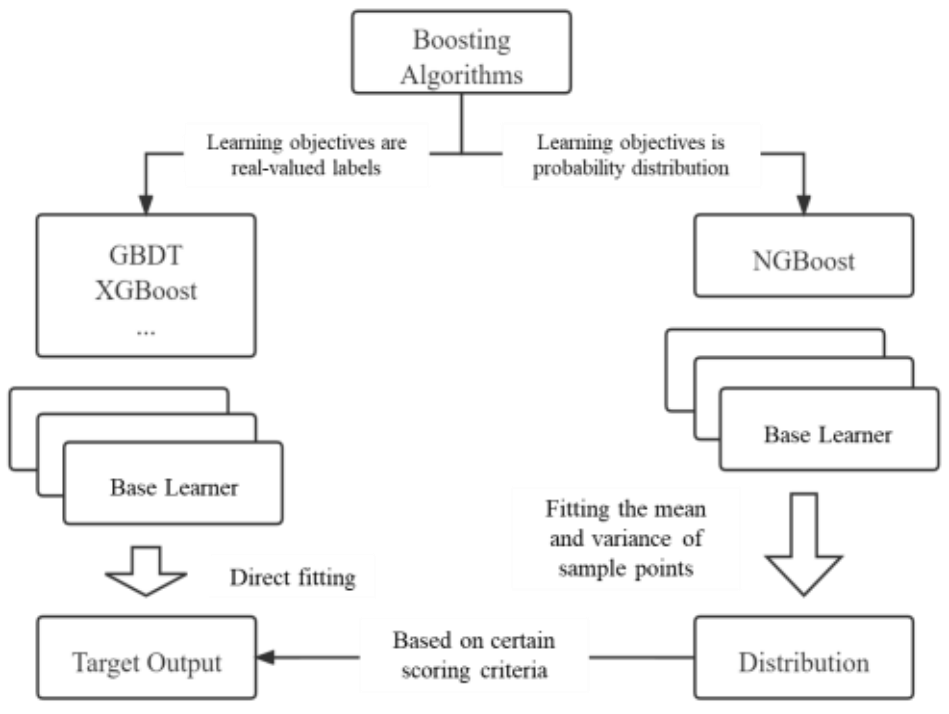

**Figure 1.** Relationship between sequential ensemble methods used in this study.

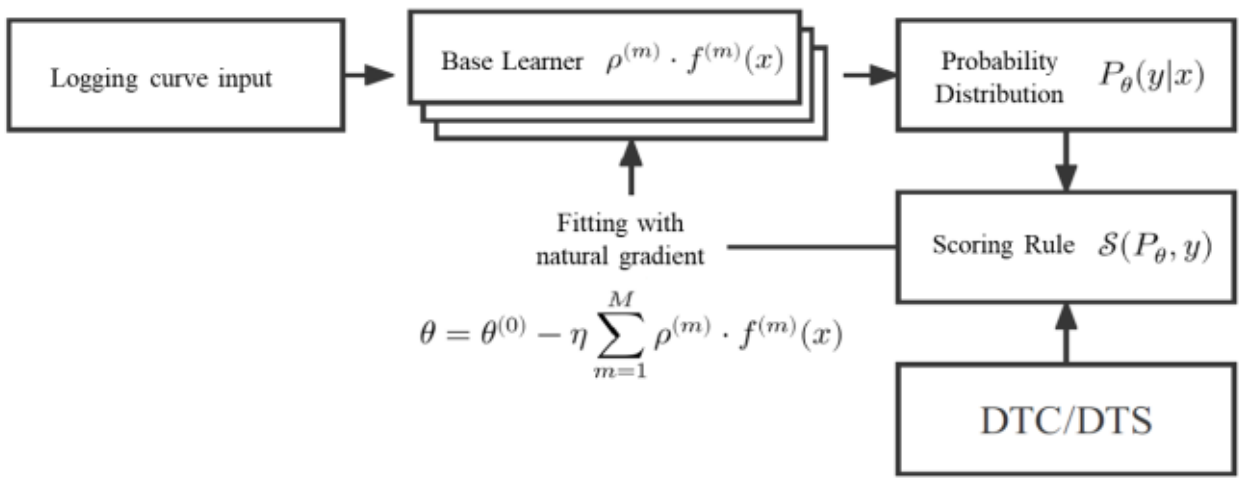

**Figure 2.** Schematic diagram of training of NGBoost algorithm for DTC/DTS prediction. The inpurt feature and DTC/DTS in the training are represented by x and y, respectively. $\boldsymbol{\theta}$, $\boldsymbol{P_\theta(y|x)}$ , $\boldsymbol{\mathcal{S}(P_\theta, y)}$ are the parameter vector of the DTC/DTS probability density distribution, the DTC/DTS probability density distribution, and the score for predictions, respectively. The base learner at the $m^{th}$ stage, $\boldsymbol{f^{(m)}(x)}$, is boosted from the input feature x and the natural gradient at the previous stage, with m=0 as the initial value.

Fig. 2 depitcs the training process of the NGBoost algorithm for DTC/DTS prediction, which comprises three main parts:

(1) Base Learner (f(x));

(2) Probability Distribution ($P_\theta$);

(3) Scoring Rule ($\mathcal{S}$).

The algorithm assumes the prediction y|x for an input feature vector x follows the probability distribution $P_\theta(y|x)$. After M stages of boosting, the algorithm obtains the probability distribution parameter $\theta$ of the prediction rather than the value given by traditional supervised learning models. The optimization objective of NGBoost is represented by Equation 1,

$$\theta = \theta^{(0)} - \eta \sum_{m=1}^{M} \rho^{(m)} \cdot f^{(m)}(x) \tag{1}$$

Here, the superscripts 0 and m represent the initial stage and the m-th stage, respectively. $f^{(m)}(x)$ is the base learner of the $m^{th}$ stage. $\eta$ and $\rho^{(m)}$ are the weighting coefficient and scaling factor of each learner, respectively.

During the training, the posterior base learner is used to fit the residual between the previous base learner and the learning target. After the training, each learner is weighted by the weight value $\rho^{(m)}$. The parameter θ is determined by the probability distribution, such as the normal distribution where the parameter θ is (μ, logσ). The loss function is constructed based on the scoring rule $\mathcal{S}$, which represents the difference between the predicted probability distribution value (e.g. mean vaule or mathematic expectation) and the label. The maximum likelihood estimation is usually used as the scoring rule (see equation 2).

$$\mathcal{S}(P_\theta, y) = -logP_\theta(y) \tag{2}$$

During the learning, the initial prediction parameters $\theta^{(0)}$ used in the model are estimated by minimizing $\mathcal{S}$ using all training labels. Then, the predicted result can be finaily obtained after M iterations according to Equation 1. Compared with Bayesian deep learning methods (Amari 1998), which can also predict the probability density distribution, the NGBoost algorithm has less computational complexity.

While complex models such as deep learning models (Jaikla et al. 2019; Li & Misra 2019; Chopra et al. 2022) have shown promising in well log prediction, lithology classification, and geological parameter prediction, there is a lack of analysis on how these models reflect the petrophysical world. The use of ensemble learning models with CART trees as base learners allows for the calculation of the global importance of features on the outputs using the Gini index ( or Gini Importance) after training. For example, Feng et al. (2021) calculated the average global importance of input logs on the predicated log in the random forest. However, the Gini index cannot provide the influence of local labels on the output, nor can it quantify the coupling relationship between different input logs, making it difficult to understand or explain the prediction process.

To addrese this issue, we apply a unified interpretive model, SHAP (Shapley

Additive Explanations) (Lundberg & Lee 2017; Lundberg et al. 2020), to help us understand the machine learning models for the DTC and DTS reconstructions. SHAP is an additive feature attribution machine learning interpretive method that represents the contribution of input features to the prediction results in each prediction. For ensemble learning models with decision trees as base learners, the SHAP model provides an interpretive method that combines the local interpretation method LIME (Local Interpretable Model-agnostic Explanations) (Ribeiro et al. 2016) and the classical Shapley value estimation method (Winter 2002). Based on the LIME method, the SHAP framework uses several linear models to approximate the complex model to be explained and calculates the Shapley importance of each input feature in these linear models one by one. This way, the contribution of each input feature can be quantitatively represented in the prediction process. For a machine learning model *f* that needs to be explained, the contribution value g(x') of the simplified input x' with M input features is defined by equation (3):

$$g(x') = \phi_0 + \sum_{j=1}^{M} \phi_j \approx f(x') \quad \left(i.e.\ \phi_0 = E_X\left(f(x)\right)\right) \tag{3}$$

Here, $\phi_0$ is the average contribution value, $\phi_j$ is the SHAP value of the $j^{th}$ input feature, f($x'$) is the model output under this input, and $E_X[f(x)]$ is the mean value of the prediction results given by the model for the entire testing set. When $\phi_0$ is set to the expected value of the model output $E_X[f(x)]$, the contribution value g(x') is approximately equal to the true output of the model f(x'). In this section, we calculate the contribution (also known as SHAP value) of different input features to the predictions using the SHAP interpretation method, taking the well-trained NGBoost model as an example.

## 3. Experimental Design and Evaluation

### 3.1 Dataset

The dataset used in this study is sourced from the 2020 PDDA (Petrophysical Data Driven Analytics) machine learning competition organized by the Society of Petrophysicists and Well Log Analysts. The model inputs consist of seven well logs: wellbore caplier (CAL), neutron porosity (CNC), natural gamma ray (GR), deep resistivity (HRD), medium resistivity (HRM), photoelectric effect (PE), and density (ZDEN). The objective of the competition was to predict compressional wave slowness (DTC) and shear wave slowness (DTS). The training and testing sets were designed in accordance with the competition background (Yu et al. 2021). Follwoing invalid value removal, the statistical results of the well logging data in the training and testing sets are presented in Tables 1 and 2, respectively. It is important to note that the study employs depth index as depth because the organizers did not provide detailed depth information. However, the depth index does not correspond to a specific depth value.

The statistical results reveal that the maximum of HRM in both datasets is significantly larger than its third quartile. This variation in resistivity logs can be attributed to the differences in resistivity across various subsurface media, where the resistivity in acquifer may be less than 0.1 Ω·m, while the resistivity of low-porosity tight formations may be as high as 1000 Ω·m. Resisitivty log tends to a normal distribution in a logarithmic coordinate. Nevertheless, abnormally high or low values may arise from measurement errors. To address this, we refer to perious studies on well logs' prediction (Misra et al. 2019; Gu et al. 2021) and replace the original values with the natural logarithm of resistivity in the training and testing sets.

**Table 1** Stastics of the logs in training set

| | **CAL (cm)** | **CNC (Φ%)** | **GR (API)** | **HRD (Ω·m)** | **HRM (Ω·m)** | **PE (Pe)** | **ZDEN (g/cm³)** | **DTC (μs/m)** | **DTS (μs/m)** |
|---|---|---|---|---|---|---|---|---|---|
| Count | 20525 | 20525 | 20525 | 20525 | 20525 | 20525 | 20525 | 20525 | 20525 |
| Mean | 8.43 | 0.27 | 49.89 | 2.60 | 5.84 | 3.83 | 2.41 | 88.31 | 182.05 |
| Std | 1.85 | 3.06 | 54.81 | 3.47 | 422.45 | 4.38 | 0.18 | 23.54 | 84.67 |
| Min | 5.93 | 0.01 | 1.04 | 0.12 | 0.13 | -0.02 | 0.68 | 49.97 | 80.58 |
| 25% | 6.63 | 0.12 | 16.04 | 0.81 | 0.80 | 0.05 | 2.24 | 70.42 | 127.15 |
| 50% | 8.58 | 0.19 | 37.50 | 1.81 | 1.83 | 3.29 | 2.47 | 79.70 | 142.68 |
| 75% | 8.67 | 0.33 | 61.14 | 3.34 | **3.46** | 7.06 | 2.56 | 102.48 | 192.76 |
| Max | 21.06 | 365.89 | 1470.25 | 206.72 | **60467.76** | 28.11 | 3.26 | 155.98 | 487.44 |

**Table 2** Stastics of the logs in testing set

| | **CAL (cm)** | **CNC (Φ%)** | **GR (API)** | **HRD (Ω·m)** | **HRM (Ω·m)** | **PE (Pe)** | **ZDEN (g/cm³)** | **DTC (μs/m)** | **DTS (μs/m)** |
|---|---|---|---|---|---|---|---|---|---|
| Count | 11088 | 11088 | 11088 | 11088 | 11088 | 11088 | 11088 | 11088 | 11088 |
| Mean | 8.63 | 0.16 | 28.97 | 4.03 | 106.75 | 7.35 | 2.48 | 76.67 | 145.35 |
| Std | 0.04 | 0.09 | 43.65 | 7.20 | 2374.62 | 1.24 | 0.15 | 14.49 | 44.39 |
| Min | 8.50 | 0.01 | 0.85 | 0.08 | 0.10 | 4.76 | 2.03 | 53.16 | 83.57 |
| 25% | 8.63 | 0.09 | 8.45 | 1.76 | 1.87 | 6.53 | 2.38 | 66.00 | 119.45 |
| 50% | 8.63 | 0.13 | 18.17 | 2.76 | 3.18 | 7.88 | 2.53 | 71.13 | 129.91 |
| 75% | 8.67 | 0.21 | 36.47 | 4.54 | **5.03** | 8.31 | 2.58 | 86.17 | 145.58 |
| Max | 8.88 | 0.56 | 1124.44 | 202.23 | **62290.77** | 13.84 | 3.02 | 126.83 | 343.95 |

## 3.2 Prediction Results and Uncertainty Analysis of Sonic Slowness

We performed cross-validation by randomly selecting 20% of the training dataset. Hyperparameters of the model, such as maximum depth of the base learner, number of estimators, and learning rate, are determined through grid search and cross-validation methods. After conducting numerous experiments, we set the maximum depth of the base learner between 2 and 8, the learning rate between 0.01 and 0.5, and the number of base learners between 50 and 500. We select the model with the best performance on the cross-validation set for testing and use four evaluation metrics, namely mean squared error, root mean squared error, explained variance score, and R2 score, to assess the performance of the logs' prediction model on both the training and testing sets. To evaluate the performance the NGBoost model's regression, we use the mean value (or mathematic expectation) at each depth point as the predicted value. Tables 3 and 4 list the best hyperparameters for DTC and DTS prediction, respectively. Tables 5-8 show the statistical results of the metrics of the ensemble learning models for DTC and DTS prediction with the best hyperparameters in this experiment.

**Table 3** Best hyperparameters for DTC prediction in the ensemble learning models

| Algorithms | Learning Rate | Max Depth | N Estimators |
|---|---|---|---|
| RF | 0.2 | 4 | 101 |
| GBDT | 0.1 | 4 | 81 |
| XGBoost | 0.2 | 4 | 302 |
| LightGBM | 0.2 | 4 | 263 |
| NGBoost | 0.04 | 4 | 489 |

**Table 4** Best hyperparameters for DTS prediction in the ensemble learning models

| Algorithms | Learning Rate | Max Depth | N Estimators |
|---|---|---|---|
| RF | 0.2 | 4 | 98 |
| GBDT | 0.1 | 4 | 51 |
| XGBoost | 0.2 | 4 | 487 |
| LightGBM | 0.3 | 4 | 254 |
| NGBoost | 0.04 | 4 | 500 |

**Table 5** Stastics of metrics of the ensemble learning models used for DTC in the training set

| Method | MSE | RMSE | MAE | EVS | R2 | Rank |
|---|---|---|---|---|---|---|
| RF | 1.053 | 1.026 | 0.549 | 0.998 | 0.998 | 1 |
| GBDT | 10.320 | 3.212 | 2.107 | 0.981 | 0.981 | 5 |
| XGBoost | 6.796 | 2.607 | 1.714 | 0.988 | 0.988 | 3 |
| LightGBM | 6.314 | 2.513 | 1.659 | 0.989 | 0.989 | 2 |
| **NGBoost** | **9.008** | **3.001** | **1.956** | **0.984** | **0.984** | **4** |

**Table 6** Stastics of metrics of the ensemble learning models used for DTC in the testing set

| Method | MSE | RMSE | MAE | EVS | R2 | Rank |
|---|---|---|---|---|---|---|
| RF | 24.000 | 4.899 | 3.128 | 0.891 | 0.886 | 5 |
| GBDT | 21.214 | 4.606 | 2.887 | 0.904 | 0.899 | 1 |
| XGBoost | 22.218 | 4.714 | 2.951 | 0.901 | 0.894 | 3 |
| LightGBM | 23.512 | 4.849 | 3.140 | 0.891 | 0.888 | 4 |
| **NGBoost** | **21.770** | **4.666** | **2.889** | **0.903** | **0.896** | **2** |

**Table 7** Stastics of metrics of the ensemble learning models used for DTS in the training set

| Method | MSE | RMSE | MAE | EVS | R2 | Rank |
|---|---|---|---|---|---|---|
| RF | 8.073 | 2.841 | 1.511 | 0.999 | 0.999 | 1 |
| GBDT | 102.726 | 10.135 | 6.795 | 0.986 | 0.986 | 5 |
| XGBoost | 51.996 | 7.211 | 4.789 | 0.993 | 0.993 | 2 |
| LightGBM | 56.295 | 7.503 | 4.852 | 0.992 | 0.992 | 3 |
| **NGBoost** | **74.859** | **8.652** | **5.738** | **0.990** | **0.990** | **4** |

**Table 8** Stastics of metrics of the ensemble learning models used for DTS in the testing set

| Method | MSE | RMSE | MAE | EVS | R2 | Rank |
|---|---|---|---|---|---|---|
| RF | 628.073 | 25.061 | 14.139 | 0.681 | 0.681 | 4 |
| GBDT | 589.512 | 24.280 | 12.100 | 0.713 | 0.701 | 3 |
| XGBoost | 571.616 | 23.908 | 11.729 | 0.722 | 0.710 | 1 |
| LightGBM | 638.666 | 25.272 | 13.833 | 0.681 | 0.676 | 5 |
| **NGBoost** | **583.929** | **24.165** | **11.769** | **0.721** | **0.704** | **2** |

Figs 3-7 illustrate the fit between the predicted and the actual results for different models.The horizontal axis is the predicted variable, while the vertical axis displays the actual value of the variable. The blue line depicts the reference line with a slope of 1, while the red line represents the fitting line of the data distribution.

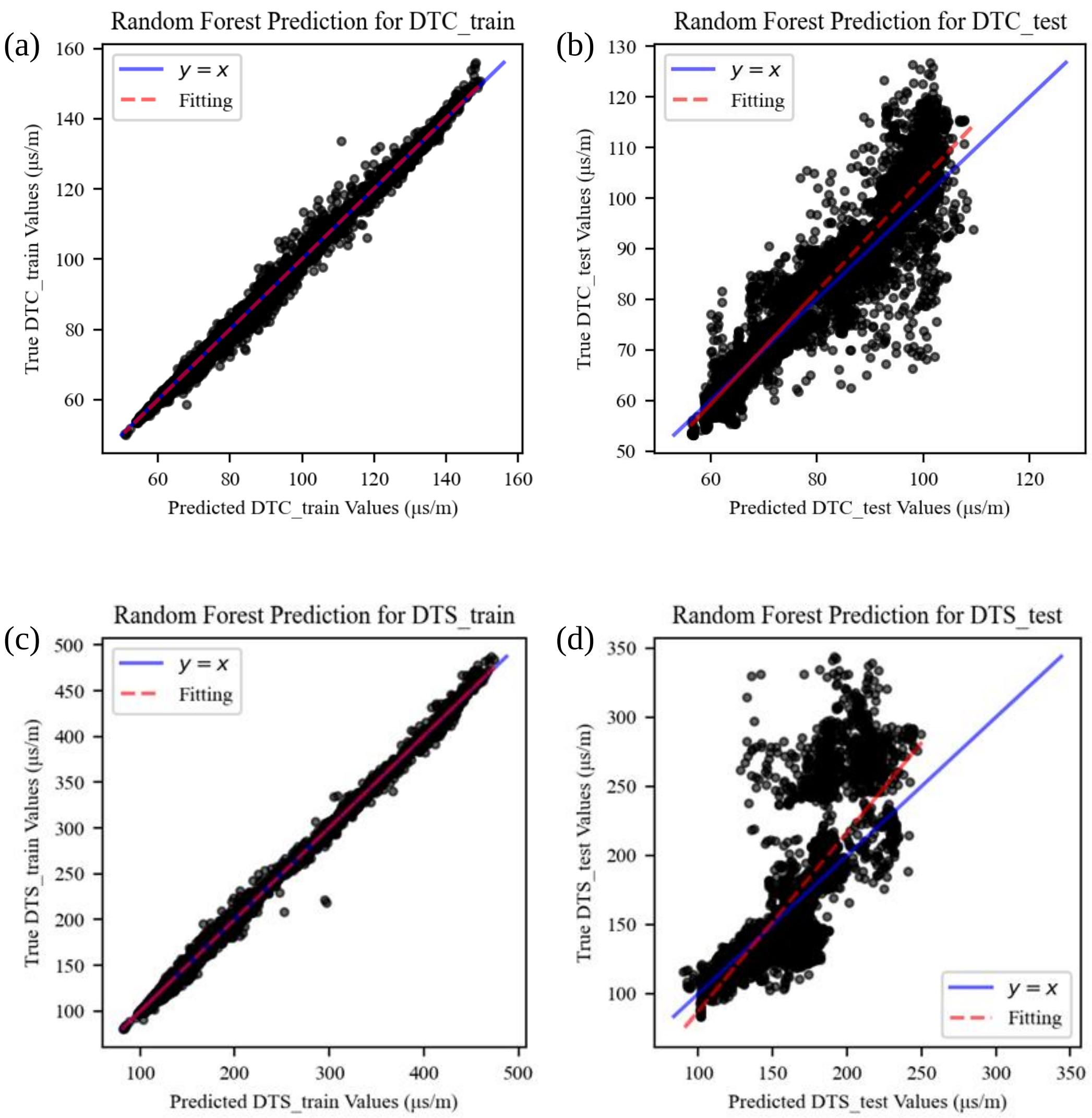

**Figure 3.** Fit between the predictions and the actual results for the training and testing sets in the random forest. (a) Traing set for DTC prediction. (b) Testing set for DTC prediction. (c) Traing set for DTS prediction. (d) Testing set for DTS prediction.

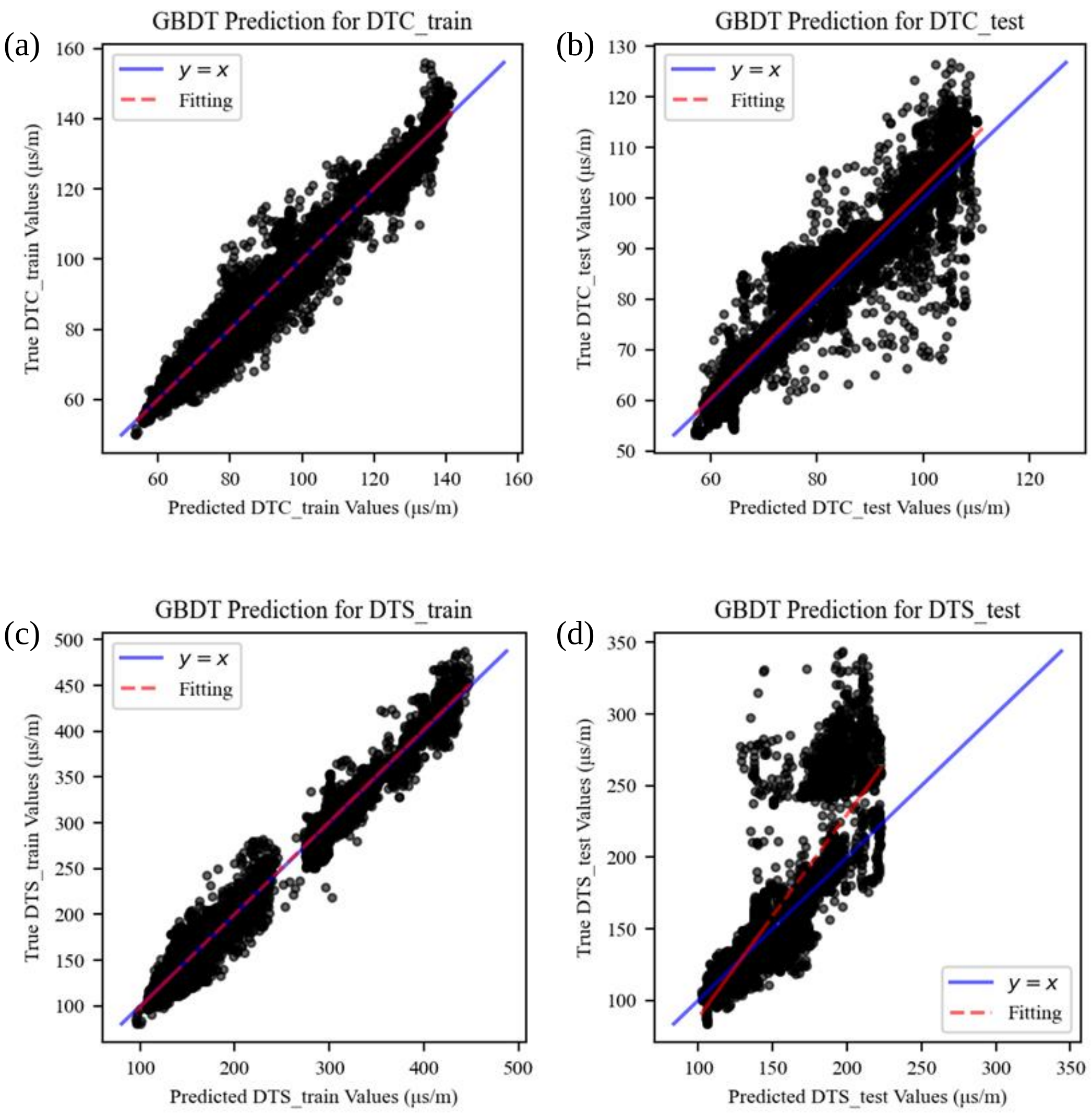

**Figure 4.** Fit between the predictions and the actual results for the training and testing sets in the GBDT. (a) Traing set for DTC prediction. (b) Testing set for DTC prediction. (c) Traing set for DTS prediction. (d) Testing set for DTS prediction.

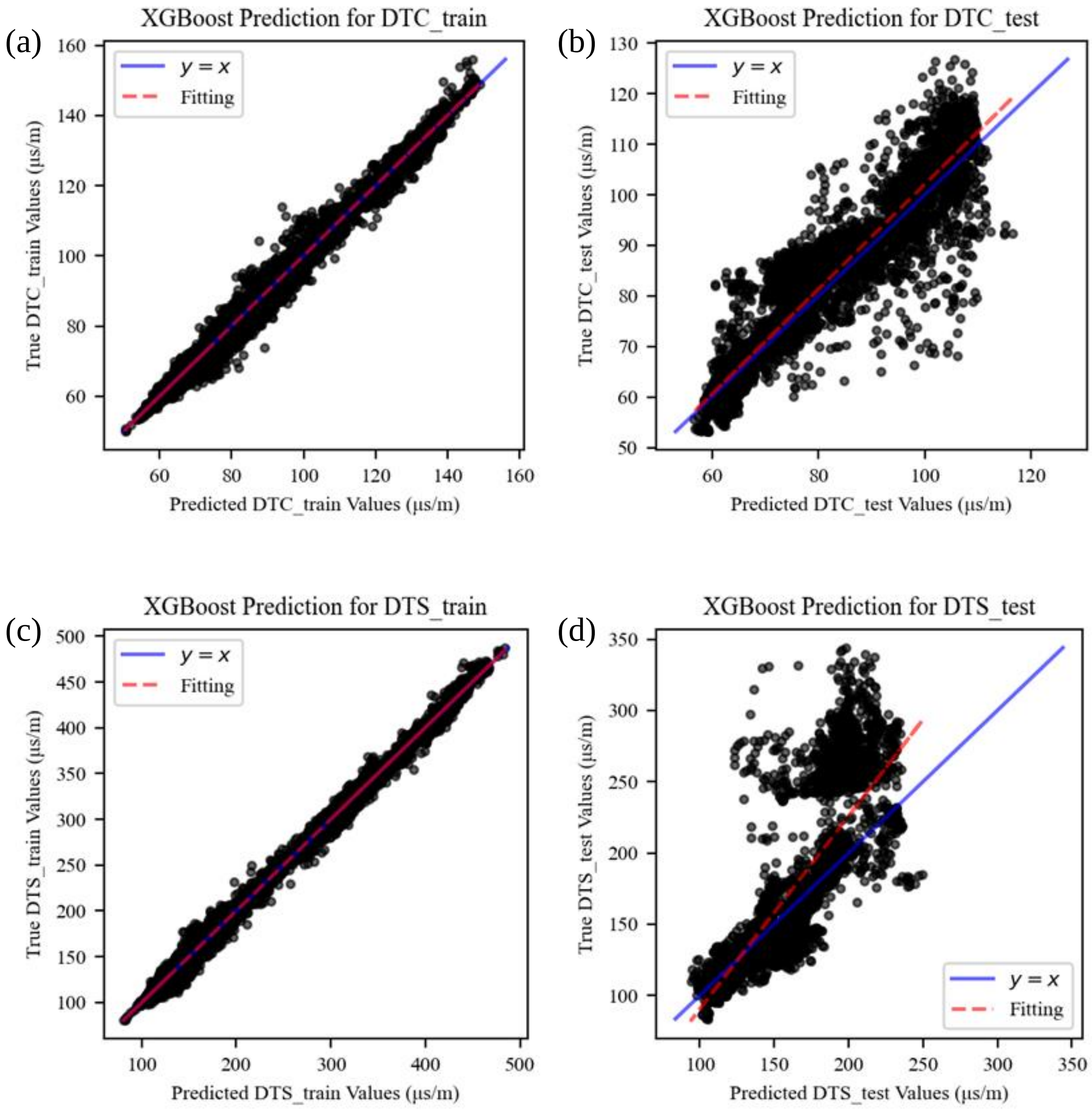

**Figure 5.** Fit between the predictions and the actual results for the training and testing sets in the XGBoost. (a) Traing set for DTC prediction. (b) Testing set for DTC prediction. (c) Traing set for DTS prediction. (d) Testing set for DTS prediction.

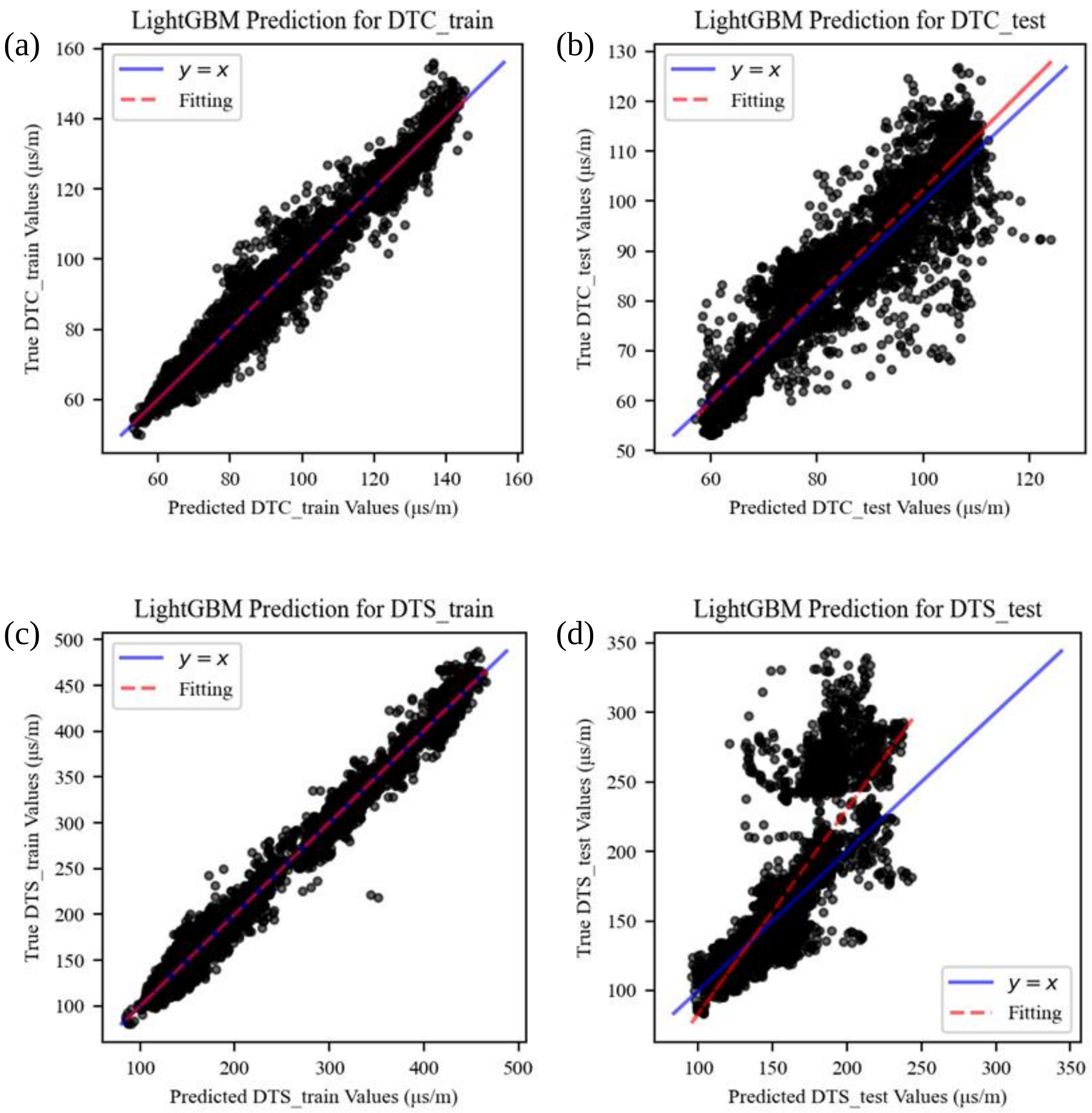

**Figure 6.** Fit between the predictions and the actual results for the training and testing sets in the LightGBM. (a) Traing set for DTC prediction. (b) Testing set for DTC prediction. (c) Traing set for DTS prediction. (d) Testing set for DTS prediction.

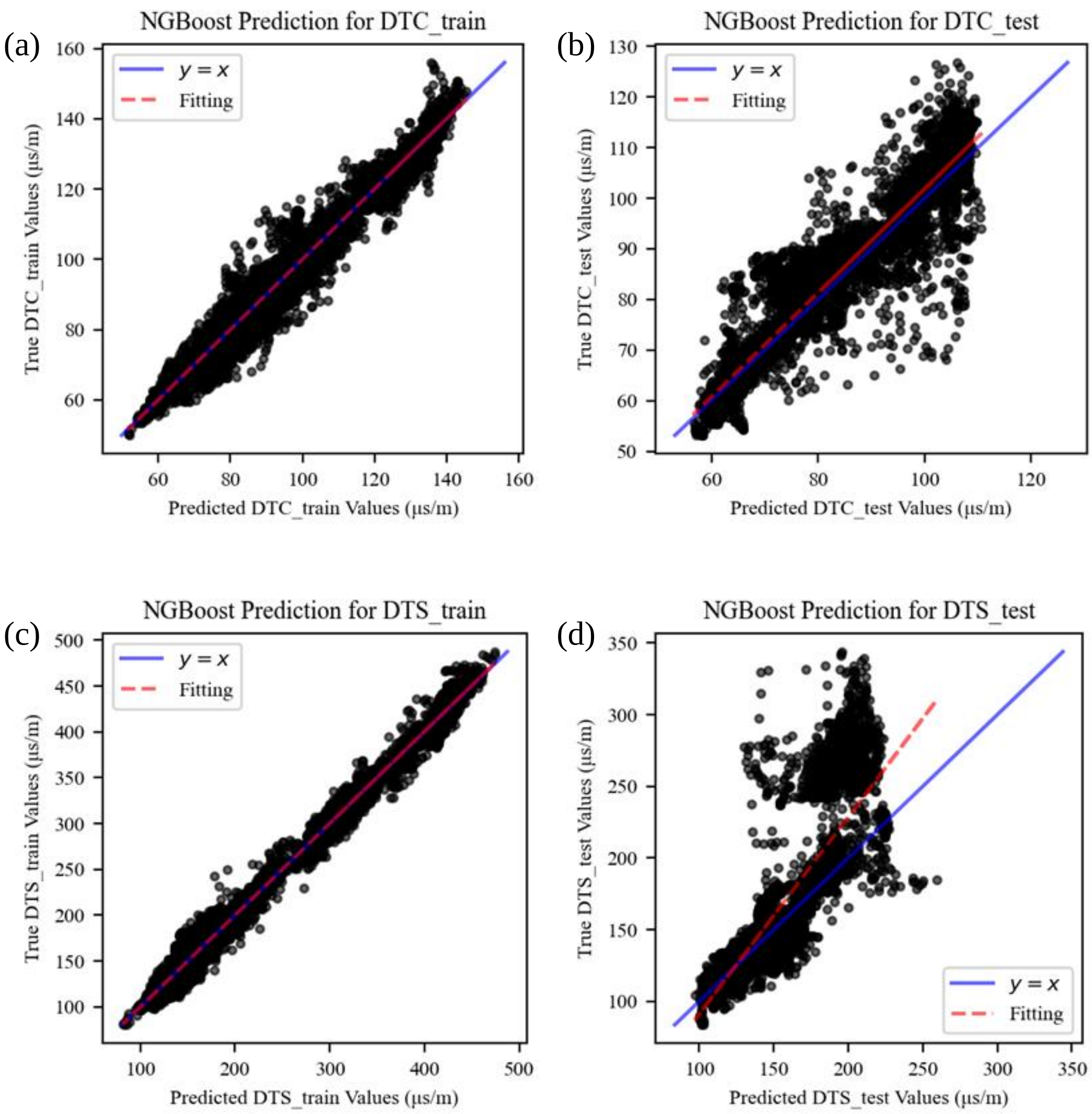

**Figure 7.** Fit between the predictions and the actual results for the training and testing sets in the NGBoost. (a) Traing set for DTC prediction. (b) Testing set for DTC prediction. (c) Traing set for DTS prediction. (d) Testing set for DTS prediction.

Based on a comprehensive analysis of the prediction and fitting performance of the machine learning models, we have drawn the following conclusions:

(1) All models exhibit good fitting performance on the training set, with predicted values close to true values. Except for the random forest model, other models have similar performances, with R2 scores higher than 0.98;

(2) The random forest model achieves the highest score on the training set but has a relatively low score on the test set. This significant difference may be due to the weak noise resistance and overfits in the training set.

(3) The NGBoost model performs well on the test set and does not overfit in the training set. This suggests that the NGBoost model predict well logs effectively without any feature processing.

**3.3 Uncertainty Analysis for the preiction**

In the field of well logging, prediction results (a certain value) are typically used as output, and the uncertainty analysis in prediction has only been studied by few researchers did some studies. For instance, Ciabarri et al. (2021) used probability classifiers based on Bayesian theory to classify lithofacies for logs and calculate the uncertainty, while Sankaranarayanan et al. (2021) and Feng et al. (2021) employed the Quantile Regression Forest to determine the uncertainty of predicted properties. Their uncertainties are derived from the voting results of all weak learners and have a greater level of randomness, which differs from scoring rules that directly obtain the mean and variance of probability distributions.

The NGBoost algorithm not only provides accurate prediction results but also calculates the probability distribution function of the prediction results using scoring rules to quantify the uncertainty of variable prediction results. Based on the probability distribution predicted by the NGBoost model, an 80% confidence interval centered on the predicted mean computed. Within this interval, the model consider an 80% likelihood that the predicted target is within it. The results obtained from the testing set indicate that approximately 78% of the observations in the DTC prediction task were within the 80% confidence interval of the model prediction results, while approximately 62% of the observations in the DTS prediction task were within the 80% confidence interval of the model prediction results. These results demonstrate that the NGBoost model has good predictive ability for the probability density of DTC and DTS.

We choose several representative depth sections to demonstrate the prediction results and their uncertainties from the NGBoost model. Fig. 8 shows the prediction results of the section with a depth index of 6000-7000. The light green dashed line represents the upper and lower boundaries of the 80% confidence interval of the prediction results, the light black dots represent the measured values, and the light blue line represents the model's predicted mean. Approximately 90% of the actual DTC values are within the 80% confidence interval of the model prediction, and approximately 70% of the actual DTS values are within the 80% interval of the model prediction. Fig. 9 shows the poor prediction results of the depth index of 1000-2000, with only about 42% of the actual DTC values within the 80% confidence interval of the model prediction and only about 19% of the actual DTS values within the 80% confidence interval of the model prediction. Through these two sets of graphs, we can find that:

(1) In sections where the NGBoost model performs well, such as Figs 8(b) and 8(d), the variance of the predictions (or standard deviation, calculated based on the probability distribution predicted for each point) is relatively stable. On the other hand, the variance of the model's predictions fluctuates greatly in sections where the model performs prooly, as shown in Figs 9(b) and 9(d).

(2) There is a certain correlation between the sections with a sudden change in the variance of the predictions and the sections with a large deviation in the prediction results. Specifically, the depth indices from 1200 to1400 and from 1600 to 1700 in Fig. 9 show significant deviations between the measured DTC and DTS values and the predicted results, and the corresponding variance prediction results also show a significant sudden change.

These results have important implications for the effective application of machine learning models in field applications. Based on the prediction mechanism of the NGBoost model, the model provides not only predictions for the target variable but also a reasonable probability distribution for each prediction, which can help petrophysicists adopt model predictions more flexibly and quantify the uncertainty in log interpretation. At the same time, the variance of the probability distribution can also serve as a validation criterion for the model's predictions. One can choose to trust the predictions of well-performing sections with low and stable variances and avoid adopting the predictions of sections with high variances.

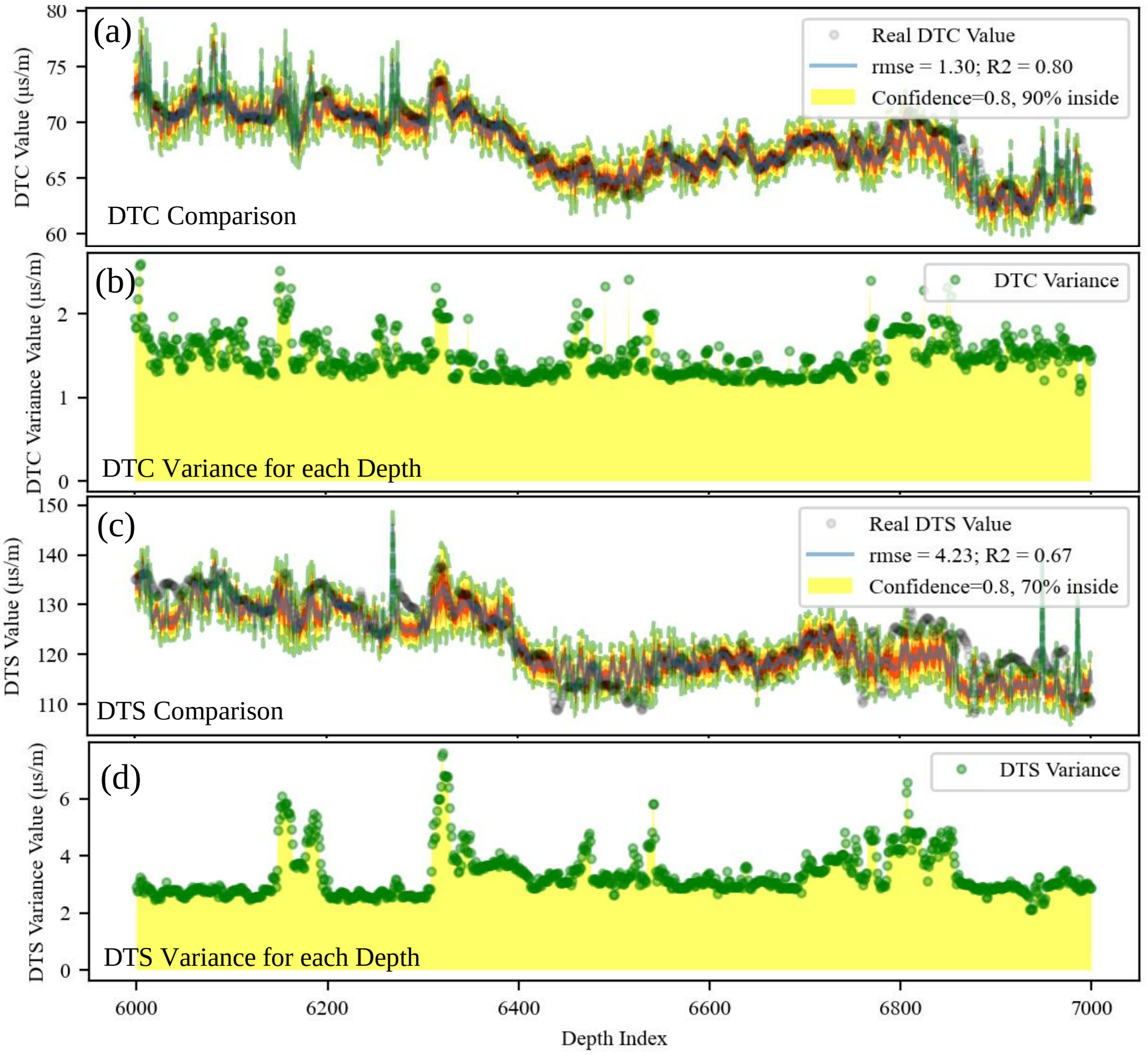

**Figure 8.** Results of sections with good prediction from the NGBoost model. (a) Probability density distribution prediction results for DTC. (b) Variance changes corresponding to each depth index from predicted DTC probability density distribution. (c) Probability density distribution prediction results for DTS. (d) Variance changes corresponding to each depth index for predicated DTS probability density distribution.

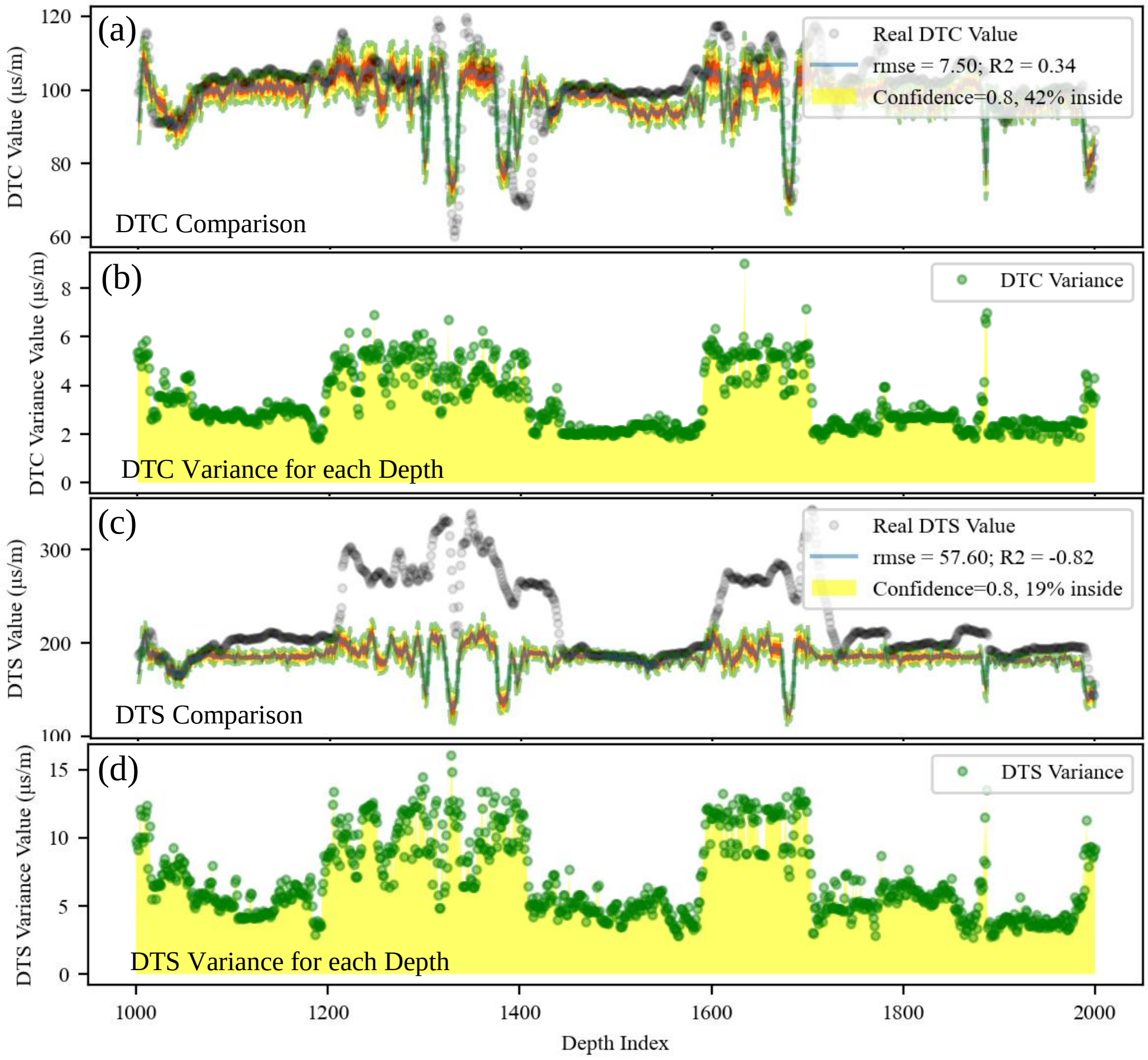

**Figure 9.** Results of sections with poor prediction from the NGBoost model. (a) Probability density distribution prediction results for DTC. (b) Variance changes corresponding to each depth index from predicted DTC probability density distribution. (c) Probability density distribution prediction results for DTS. (d) Variance changes corresponding to each depth index for predicated DTS probability density distribution.

### 3.4 Interpretability analysis of the prediction model

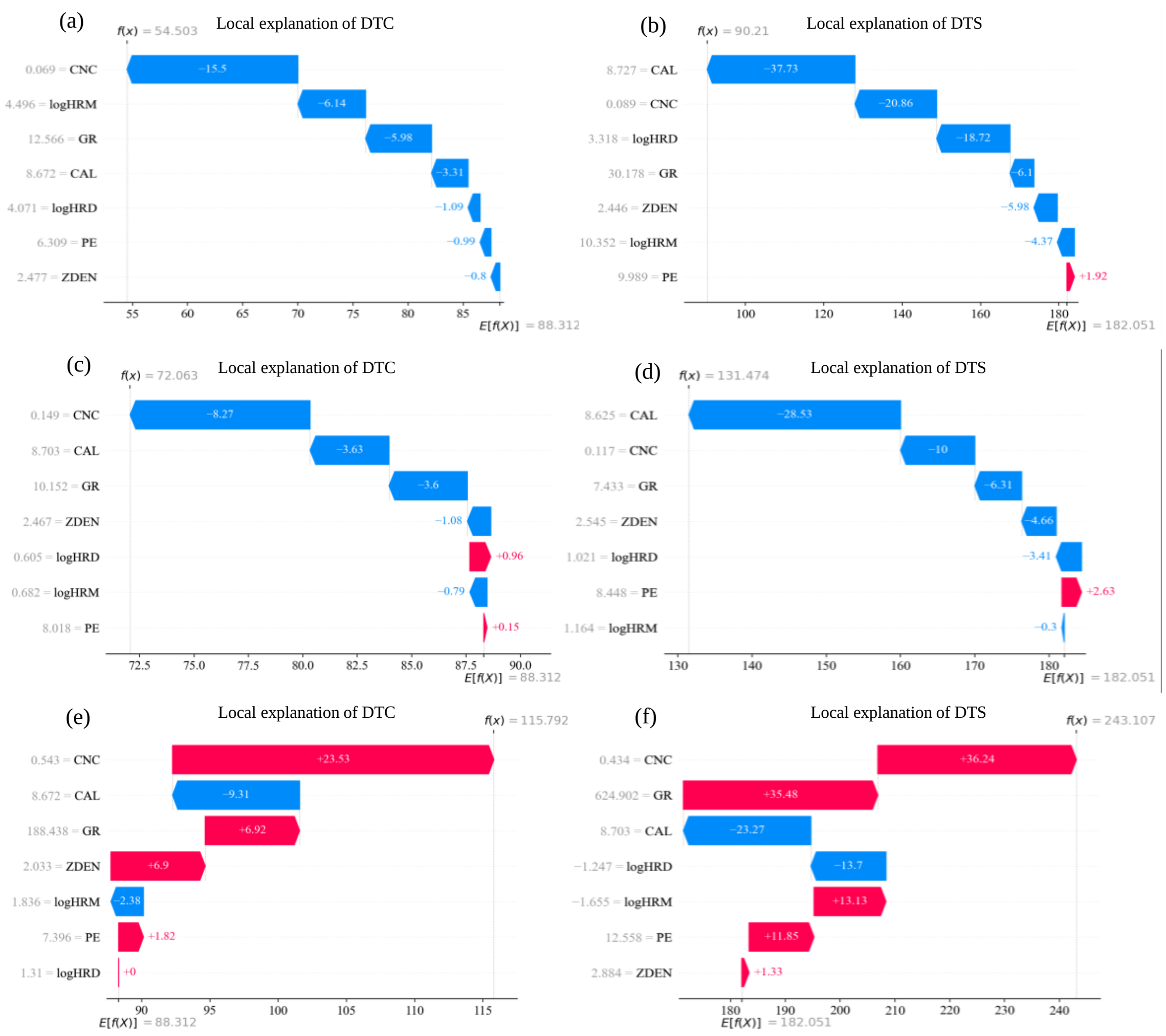

**Figure 10.** Preliminary interpretation for the contributions of the input logs on the predicated DTC and DTS probability distribution functions in the NGBoost model. (a) Minimum predicted DTC value. (b) Minimum predicted DTS value. (c) Median value of predicted DTC. (d) Median value of predicted DTS. (e) Maximum predicted DTC value. (f) Maximum predicted DTS value.

In this section, we will explain the NGBoost prediction model using SHAP value. Fig. 10 shows the model's prediction results of the minimum, median, and maximum values for the predicted of DTC and DTS (the mean value of the probability distribution function) to demonstrate each input log's contribution to the prediction results. The arrow's width represents each log's contribution to the SHAP value of the prediction (i.e., $\varphi_j$ in Equation 3), with blue indicating negative values (negative contributions) and red indicating positive values (positive contributions). In the figure, f(x) is the output at the depth index, and E[f(x)] is the mean of the model output on the entire test set. The left gray number of each log variable represents its log value at the depth. The y-axis is arranged from largest to smallest absolute value according to the feature contribution. The following observations can be made:

(1) The neutron porosity CNC has a significant contribution (larger absolute value of SHAP value) to the predictions of DTC and DTS, but the contribution varies.

(2) When the predicted slowness is small (Figs 10a and 10b), almost all the input logs have negative contributions to the prediction results. When the predicted slowness is large (Figs 10e and 10f), the CNC and the GR have a relatively large positive contribution to the model's prediction results.

By combining the feature values corresponding to CNC and GR, we find that a higher CNC and GR make the model more inclined to give a larger slowness prediction, which is consistent with the petrophysical model's understanding.

We further analyze the the impact of different inputs on DTC and DTS (see Figs 11 and 12). Figs 11(a) and 12(a) show the average feature importance of the prediction results, with the y-axis representing different input logs and the blue strip's length representing the average importance of each log, i.e., the mean of all SHAP values' absolute values. Figs 11(b) and 12(b) show the global feature importance of the prediction results, with the y-axis representing different input logs, and each point in the input log representing different data records' corresponding SHAP values. The feature value are color-coded with dark representing large feature values and light representing small feature values, which are arranged from left to right based on their SHAP value, and data records with the same SHAP value are stacked along the y-axis.

The figures reflect the following information:

(1) The neutron porosity has a significant impact on the predictions of DTC and DTS, while the PE has a relatively small impact on the predictions. Borehole caplier has a significant impact on the predictions of DTC and DTS, with a larger impact on DTS. This is consistent with petrophysics because varations in borehole diameter can cause significant fluctuations in sonic slowness. The machine learning model captured the complex influence of borehole diameter on sonic slowness. Furthermore, it has a relatively high importance in the machine learning model, regardless of the borehole diameter amount, which is needed to futher analyze.

(2) CNC and GR have a clear positive correlation with the model output as the feature value increases from negative to positive, while the density (ZDEN) and logarithmic of the medium resistivity (logHRM) have a significant negative correlation with the model output as the feature value increases. According to petrophysical knowledge, mudstone has a higher DTC, while tight sandstone, limestone, and dolomite have a lower DTC and DTS. Mudstone contains more bound water and has a relatively higher hydrogen index, corresponding to a higher CNC. The GR increases with the increase of mud content, while tight sandstone and limestone have lower porosity, higher density, and larger resistivity. The model assumes that higher CNC and GR will increase the predicted results of DTC and DTS, while higher ZDEN and logHRM will decrease the predicted results of DTC and DTS. This indicates that we can explain the black-boxed machin learning model well using SHAP. Furthermore, the explanation conforms to general petrophysical experience for the related features, indicating the effectiveness of the machine learning model in this study.

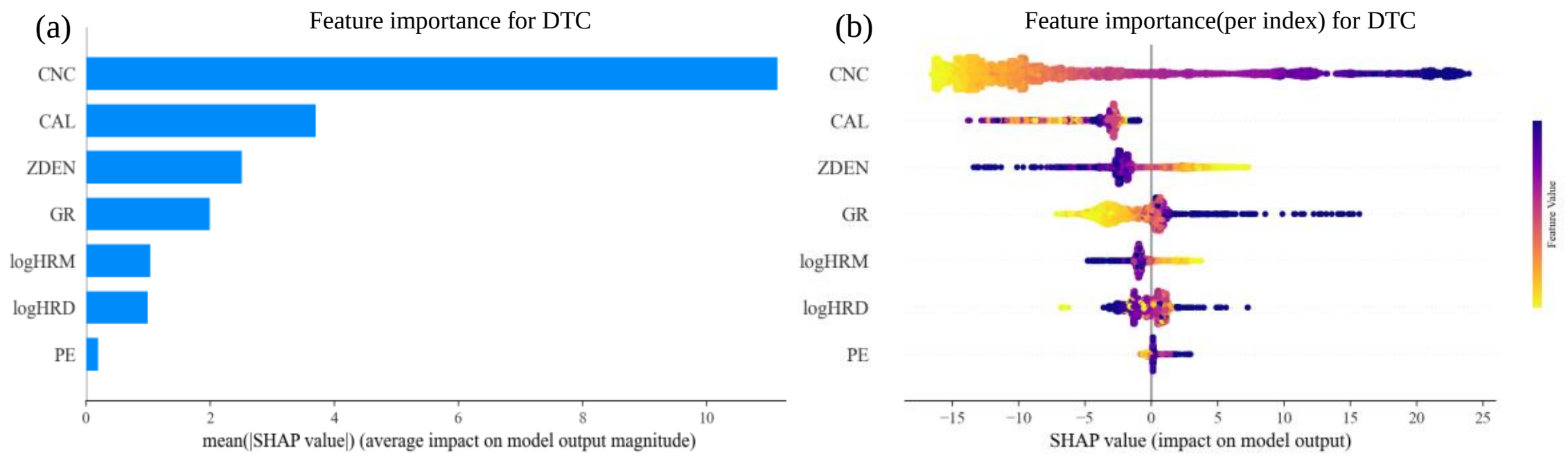

**Figure 11.** Statistical analysis of feature importance for DTC prediction results. (a) Average feature importance for DTC prediction results. (b) Global feature importance for DTC prediction results.

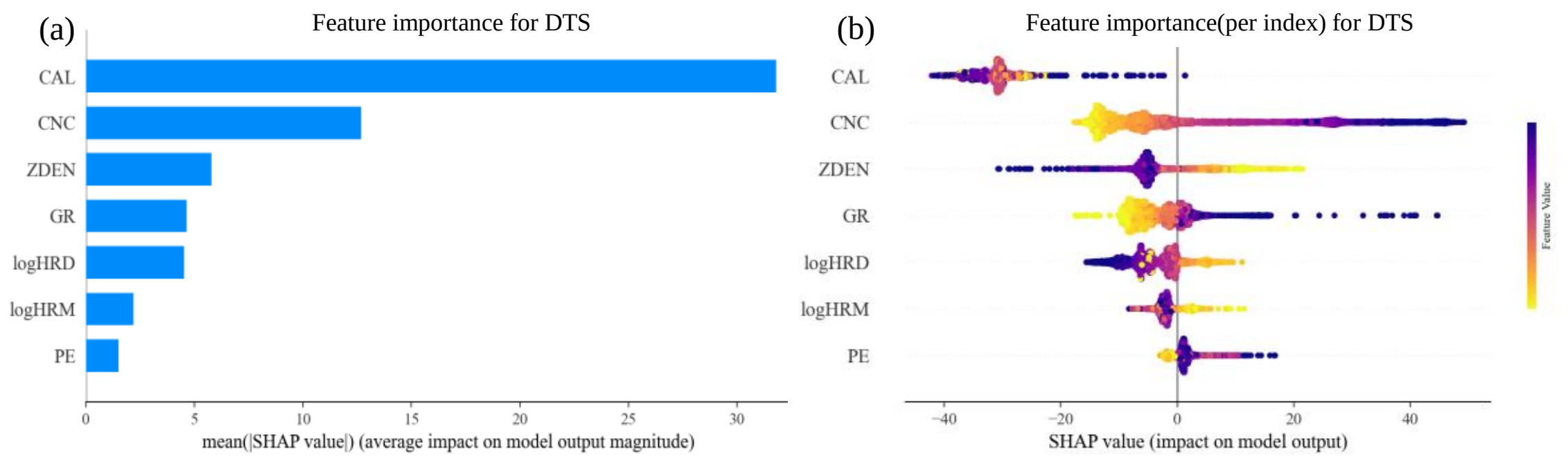

**Figure 12.** Statistical analysis of feature importance for DTC prediction results. (a) Average feature importance for DTS prediction results. (b) Global feature importance for DTS prediction results.

Except for the impact of the input logs on the outup, there is a coupling relationship among input logs where the feature importance of a certain input log may be related to other input logs. In this study, we investigate the coupling relationship among input logs in the machine learning model for predicting DTC① using scatter plots as shown in Figs 13 to 15. Each data point in the scatter plots is obtained from a certain depth index. For

① The importance of these three features in DTS prediction is very similar to that of DTC. We will not further demonstrate this due to space limitations.

example, we use the borehole caplier (CAL) as horizontal axis in Fig. 13 to investigate the coupling relationship between CAL and other input logs where CNC, ZDEN, logHRM, logHRD, GR, and PE are the vertical axes in the subplots from Figs 13(a) to 13(f). The SHAP values of the CAL on the DTC prediction are color-coded (as shown in the colorbar). Our findings show that the importance of borehole caplier (CAL) has a wide-area distribution with other logs (Fig. 13) where for a certain CAL, its importance ranges from -13 to -1 when the logs on the vertical axises change.

In Fig. 13(a), the importance of CAL on the predicted DTC becomes more negative as CNC increases (Fig. 13a). Combining with the results in section 3.3, we find that CNC has a higher average feature importance than other logs on the DTC pretication, and its importance increases positively with its value increasing. While there is no obvious relationship between the CAL value and its feature importance. In detail, for different points (different depth) in the scatter plots having a same CAL, the importance of the CAL on the predicted DTC decreased significantly with the increase of CNC (the absolute value of SHAP value increases, but the SHAP value is negative). Since varations in wellbore diameter can cause errors in CNC log, petrophysicists usually correct the impact of borehole diameter on CNC log before further interpretation in conventional log interpretation. In the machine learning model, we did not perform borehole diameter correction on the CNC in advance, but the machine learning model assigned a relatively high importance to the influence of the CAL, achieving the effect of borehole diameter correction in conventional logging interpretation. This gives us an answer as why CAL has a relatively high importance in the machine learning model, regardless of the size of borehole diameter. In addition, the lack of significant coupling relationship between CAL and other logs during prediction is also consistent with petrophysical perception (Figs 13b to 13f).

According to the experience from petrophsycics, we know that the CNC should be important on the sonic slowness preidiction. We further study the coupling relationship between CNC and other input logs in Fig. 14. We find that the importance of CNC on the predicted DTC is positively correlated with CNC value, that is, the larger the CNC value, the greater its SHAP value, and the greater its positive impact on the machine learning model. When the CNC exceeds 0.35, its SHAP value of CNC no longer increases. Combining with the content of the previous section, a large CNC corresponds to a large GR value, generally reflecting a higher mud content in the formations which have a large slowness. Therefore, the positive correlation between CNC and GR is consistent with the analysis in the beginning of this section (Fig. 14e). While the tight sandstone, limestone and dolomite which have lower DTC and DTS have lower CNC, higher ZDEN, HRM and HRD. Thus, CNC shows negative correlation with ZDEN, HRM and HRD in generally (Figs 14b, 14c, 14d).

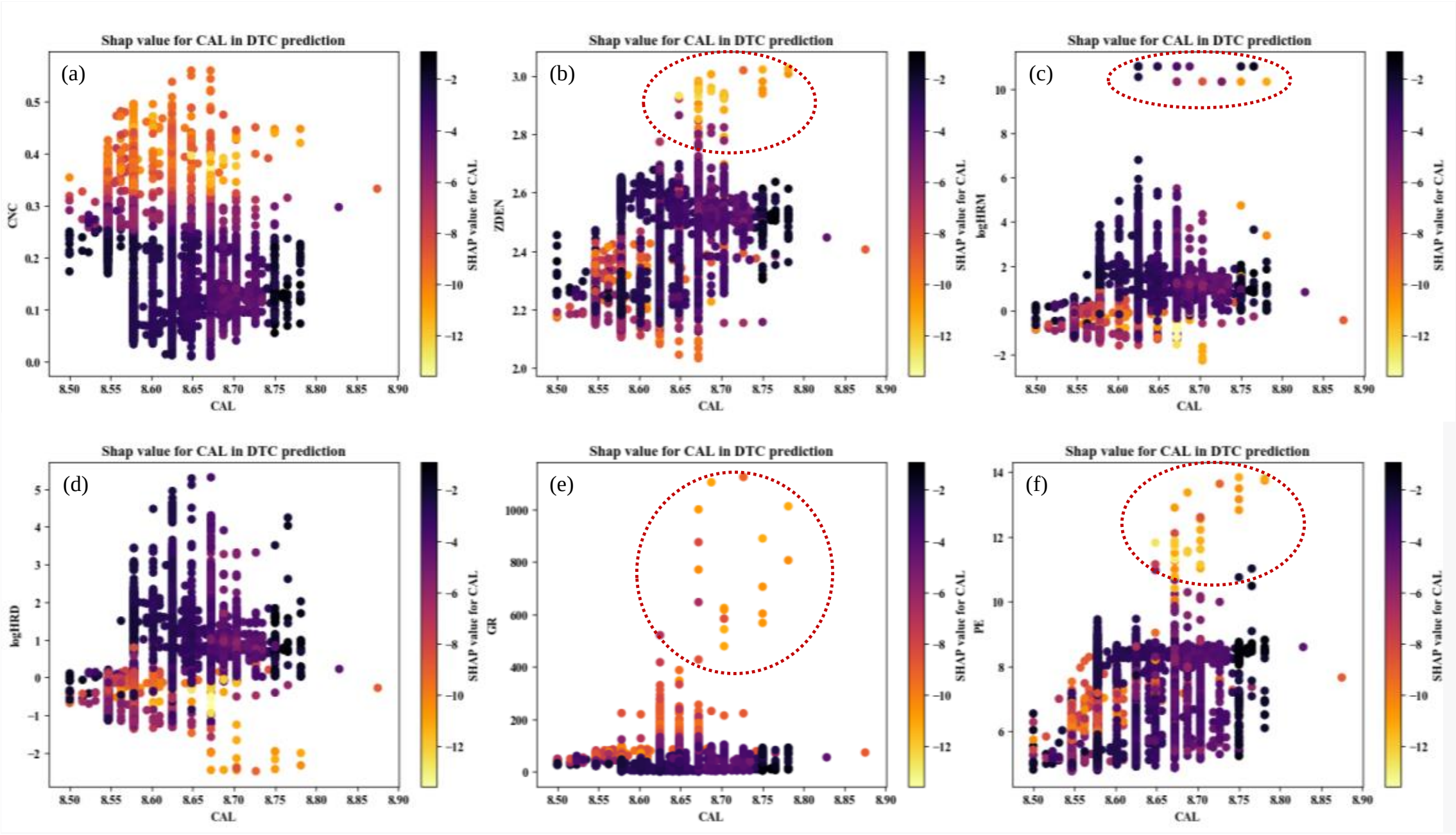

**Figure 13.** Relationship between the feature importance of wellbore diameter (CAL) and other features in DTC prediction. Outliers are marked with dashed-circles. (a) Relationship between CAL and CNC. (b) Relationship between CAL and ZDEN. (c) Relationship between CAL and logHRM. (d) Relationship between CAL and logHRD. (e) Relationship between CAL and GR. (f) Relationship between CAL and PE.

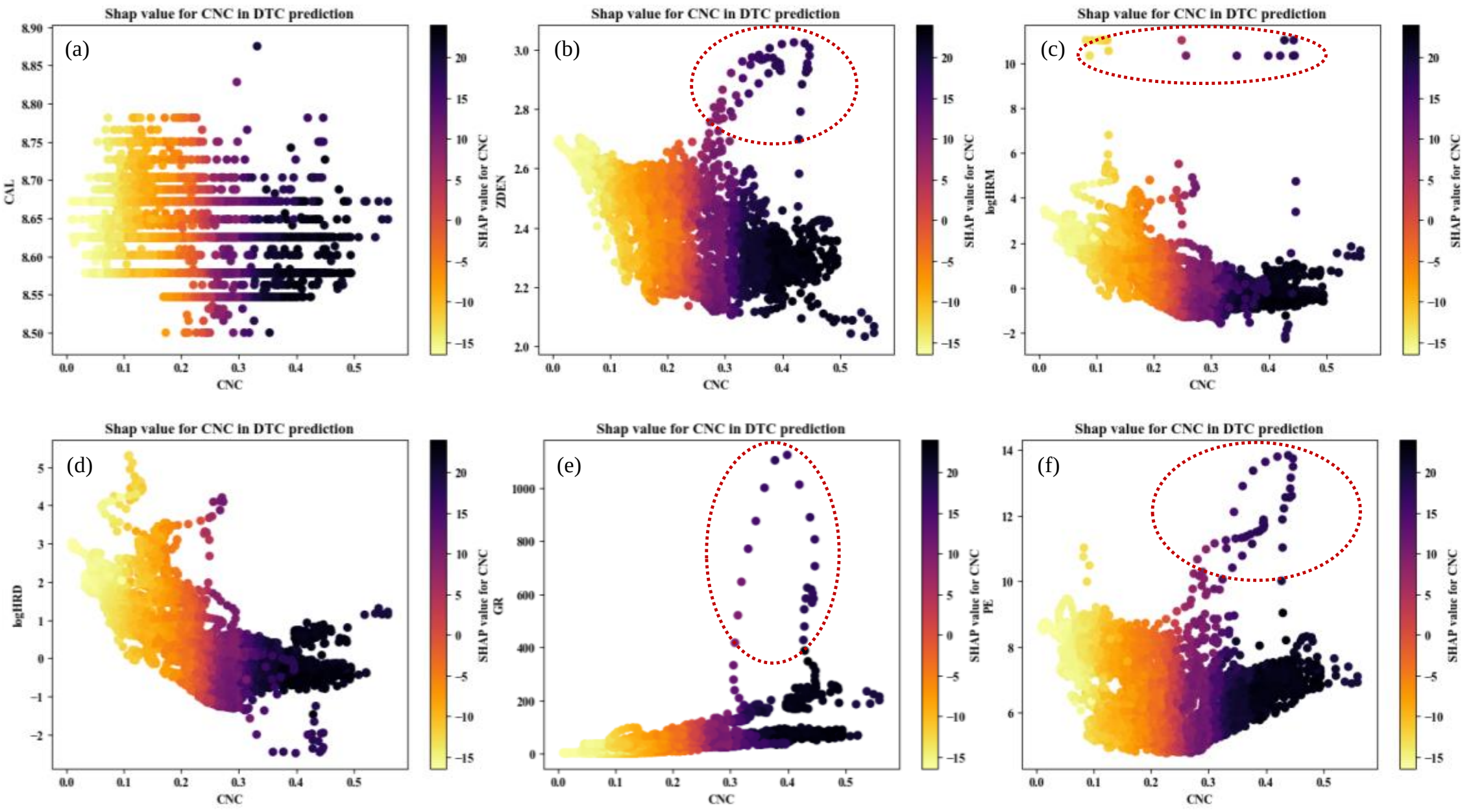

**Figure 14.** Relationship between the feature importance of CNC and other features in DTC prediction. Outliers are marked with dashed-circles. (a) Relationship between CNC and CAL. (b) Relationship between CNC and ZDEN. (c) Relationship between CNC and logHRM. (d) Relationship between CNC and logHRD. (e) Relationship between CNC and GR. (f) Relationship between CNC and PE.

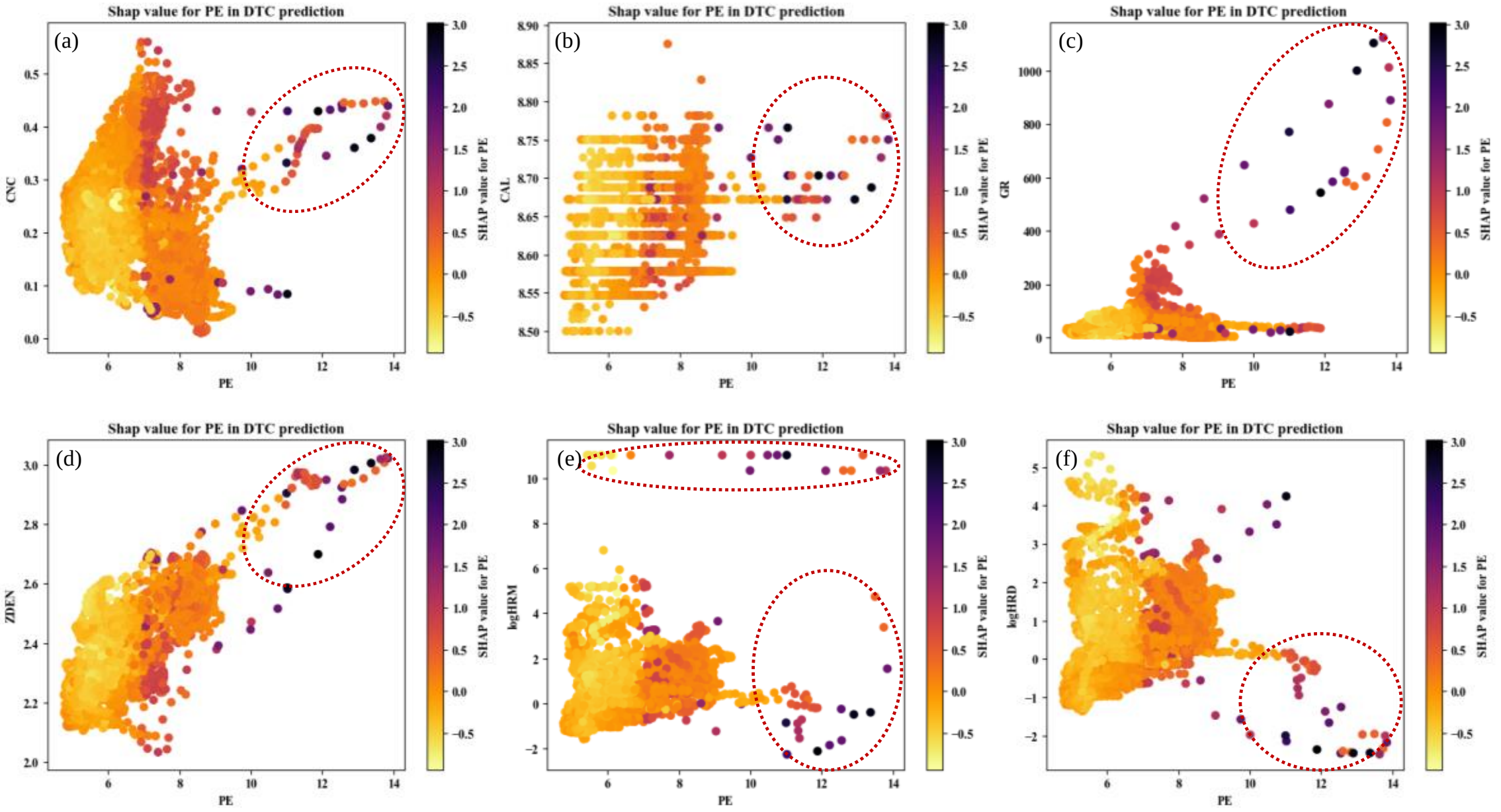

**Figure 15.** Relationship between the feature importance of PE and other features in DTC prediction. Outliers are marked with dashed-circles. (a) Relationship between PE and CNC. (b) Relationship between PE and CAL. (c) Relationship between PE and GR. (d) Relationship between PE and ZDEN. (e) Relationship between PE and logHRM. (f) Relationship between PE and logHRD.

As for the PE, a log of photoelectric properties, is defined as $(Z/10)^{3.6}$ where Z is the average atomic number of the formation, which is relatively sensitive to lithology. From the perspective of petrophysical models, PE values are higher for formations with lower mud content (e.g., dolomite and limestone) and those with higher mud content (e.g. clay), while relatively lower for sandstone, which are generally below 10. Formations with PE values higher than 10 may contain rare heavy minerals. Therefore, there is a weak correlation between the input PE log and DTC and DTS logs. We can see that most of the SHAP values of PE in Fig. 15 are concentrated between -1 and 1 which futher illustrates that the weak influence of PE on DTC and DTS predications. The results of the model interpretation in Fig. 15 show that the coupling relationship between PE and other features is complicated. When the PE value is greater than 10, its SHAP value has a large variation, indicating that the lithology may change greatly, and thus has a greater impact on DTC and DTS. However, there is no obvious correspondence between PE and any of the other characteristics, and most of the feature importance is small. The interpretation results are consistent with the petrophsycial knowledge that the correlation between PE and DTC and DTS are poor.

We note that there are some obvious outliers (e.g. GR > 300, PE > 10, etc.) marked with dashed-circles in Figs 13 to 15, which may be data anomalies caused by some external environmental influences during the logging operation, or obvious lithology changing in some deep intervals. Here we give two exmaples to analyze the outliers in order to not only deepen the understanding of the lithology of geological intervals, but also guide the further data screening and cleaning, or get better prediction results.

In Fig. 16, we show the logs from depth index point 9420 to 9445 where CAL, GR and PE increase, while HRD decreases, and HRM fluctuates greatly. Borehole maybe collapsed in this section according the data analysis and this results in abnormal GR, HRM, and HRD logs. Fig. 17 shows an example for the depth indexes from 9760 to 9790 with changing lithology. Because the ZDEN log of this section increases and PE values are greater than 10, the formation in this section would be tight formation containing heavy minerals.

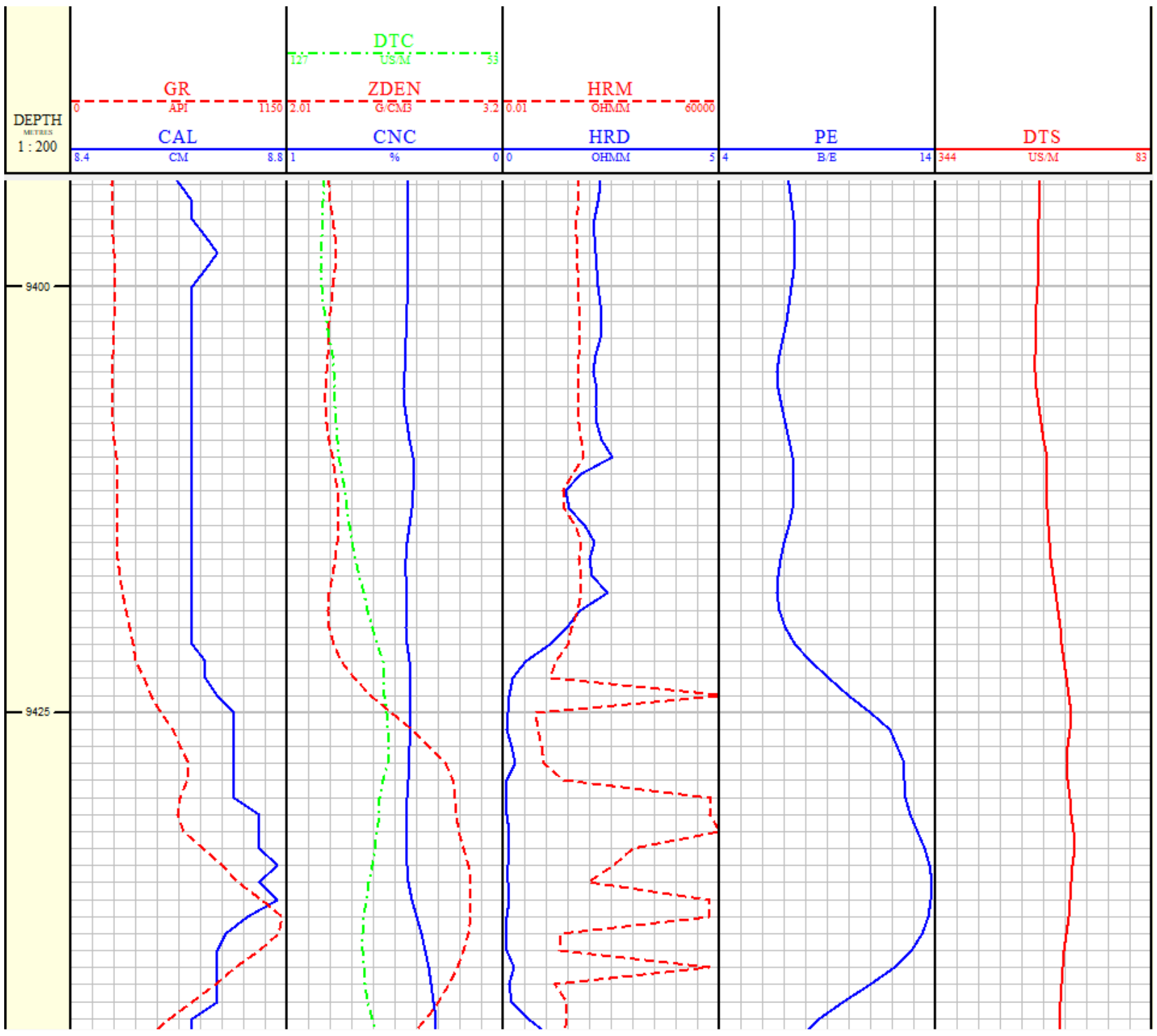

**Figure 16.** Logs for the depth index from 9420 to 9445.

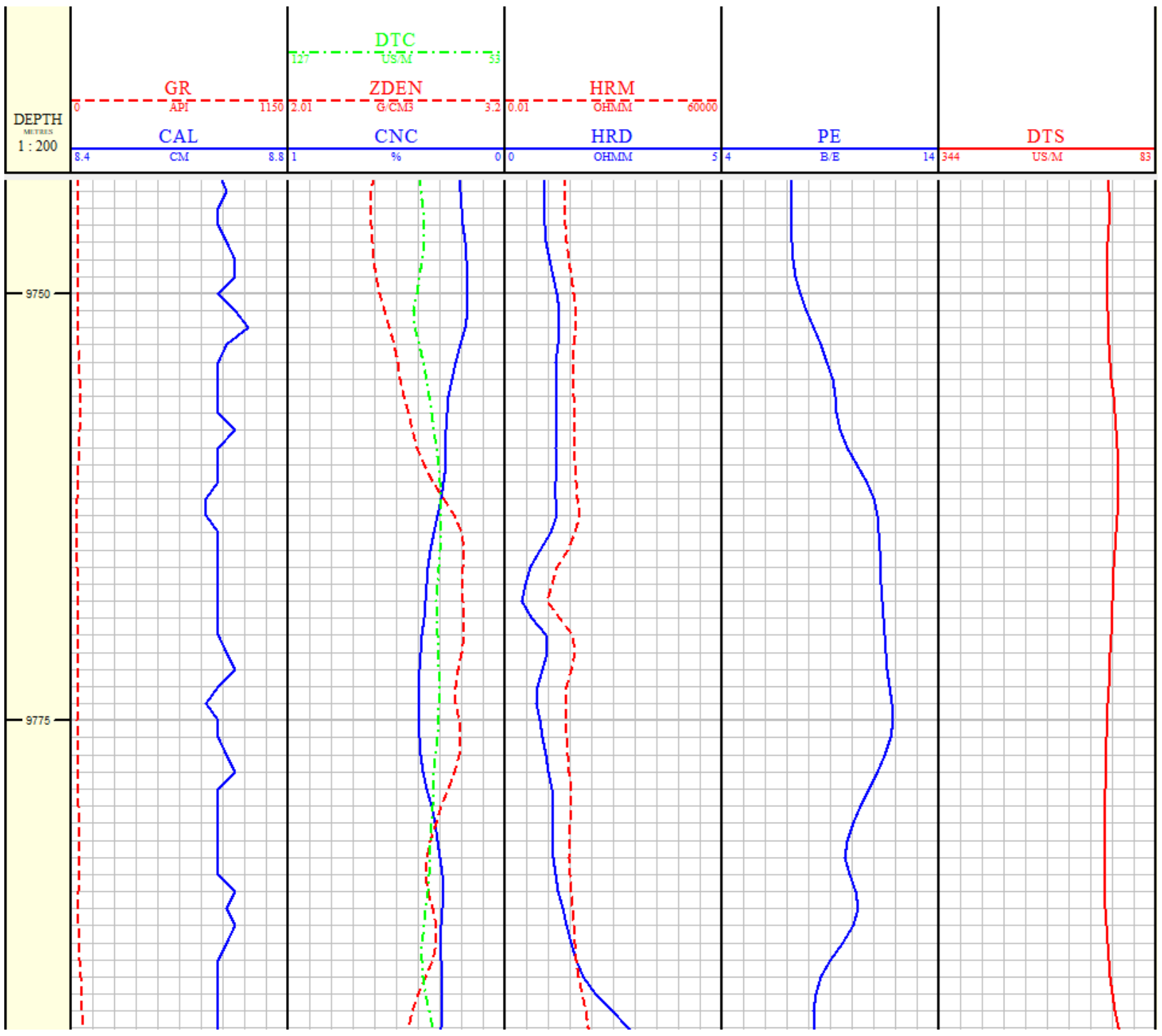

**Figure 17.** Logs for the depth index from 9745 to 9790.

## 4. Conclusions

The study compared the performance of five different ensemble learning method- Random Forest, GBDT, XGBoost, LightGBM, and NGBoost- in reconstructing logs on the training and testing sets. Except for Random Forest, which had a lower score due to overfitting caused by weak noise resistance, the other models showed relatively close predictive performance on the training set. NGBoost outperformed most other methods on the testing set and did not overfit in the training set. Moverover, NGBoost can provide reasonable probability distributions for each prediction, enabling quantitative analysis of the uncertainty and quality of the reconstructed logs. The importance of each input log to the predictive performance of the NGBoost model was calculated based on the SHAP interpretation model, and the coupling relationship between each input log was evaluated to interpret the model results.

Based on the study, the following suggestions and understandings were obtained:

(1) When using the NGBoost algorithm for log reconstruction, the variance value in the probability distribution can be used to determine the quality of the prediction results. Larger variance values can indicate problematic sections, which can effectively carry out logging interpretation tasks. This suggestion is significance for deploying machine learning models in practical logging applications.

(2) The machine learning model's predictive results showed that higher neutron porosity and GR make the model give larger sonic slowness, consistent with petrophysics models.

(3) The machine learning model can capture the complex influence of borehole diameter on sonic slowness.

(4) Although petrophysicists usually correct the impact of borehole diameter on CNC log before interpretation in traditional log interpretation, the machine learning model achieves borehole diameter correction by assigning a relatively high importance to the CAL, even without performing borehole diameter correction on the CNC in advance.

**Acknowledgements**

This study is supported by National Natural Science Foundation of China (Grant Numbers: 41974150 and 42174158), a Supporting Program for Outstanding Talent of the University of Electronic Science and Technology of China (No. 2019-QR-01), and a Project of Basic Scientific Research Operating Expenses of Central Universities (ZYGX2019J071 and ZYGX2020J013).

## Data Availability Statement

The dataset used in this study is sourced from the 2020 PDDA (Petrophysical Data Driven Analytics) machine learning competition organized by the Society of Petrophysicists and Well Log Analysts (https://github.com/pddasig/Machine-Learning-Competition-2020).